\pdfoutput=1

\documentclass[11pt]{article}

\usepackage{acl}

\usepackage{times}
\usepackage{latexsym}
\usepackage{caption}
\usepackage{graphicx}
\usepackage{multirow}
\usepackage{graphics}
\usepackage{xcolor}

\usepackage[T1]{fontenc}

\usepackage[utf8]{inputenc}

\usepackage{microtype}

\usepackage{inconsolata}

\usepackage{booktabs}

\title{Advancing State of the Art in Language Modeling}

\author{David Herel \\
  FEE, CTU \\
  CIIRC CTU \\
  \texttt{david.herel@seznam.cz} \\\And
  Tomas Mikolov \\
  CIIRC CTU \\
  \texttt{tmikolov@gmail.com} \\}

\begin{document}
\maketitle
\begin{abstract}
Generalization is arguably the most important goal of statistical language modeling research.
Publicly available benchmarks and papers published with an open-source code have been critical to advancing the field.
However, it is often very difficult, and sometimes even impossible, to reproduce the results fully as reported in publications.
In this paper, we propose a simple framework that should help advance the state of the art in language modeling in terms of generalization. 
We propose to publish not just the code, but also probabilities on dev and test sets with future publications so that one can easily add the new model into an ensemble.
This has crucial advantages: it is much easier to determine whether a newly proposed model is actually complementary to the current baseline. Therefore, instead of inventing new names for the old tricks, the scientific community can advance faster. Finally, this approach promotes diversity of ideas: one does not need to create an individual model that is the new state of the art to attract attention; it will be sufficient to develop a new model that learns patterns which other models do not. Thus, even a suboptimal model can be found to have value. Remarkably, our approach has yielded new state-of-the-art results across various language modeling benchmarks up to 10\%.
\end{abstract}

\section{Introduction}

Statistical language modeling has been around since the work of Claude Shannon in 1950 \citep{shannon1951prediction}. For decades, it was believed that the state of the art in language modeling are the n-gram techniques. While dozens of more advanced approaches have been developed (including neural language models), it was believed that those alternatives only work on small datasets, and with increasing amount of data, the improvements over n-grams vanish \citep{goodman}.

\begin{figure}[h]
\centering
\includegraphics[width=\linewidth]{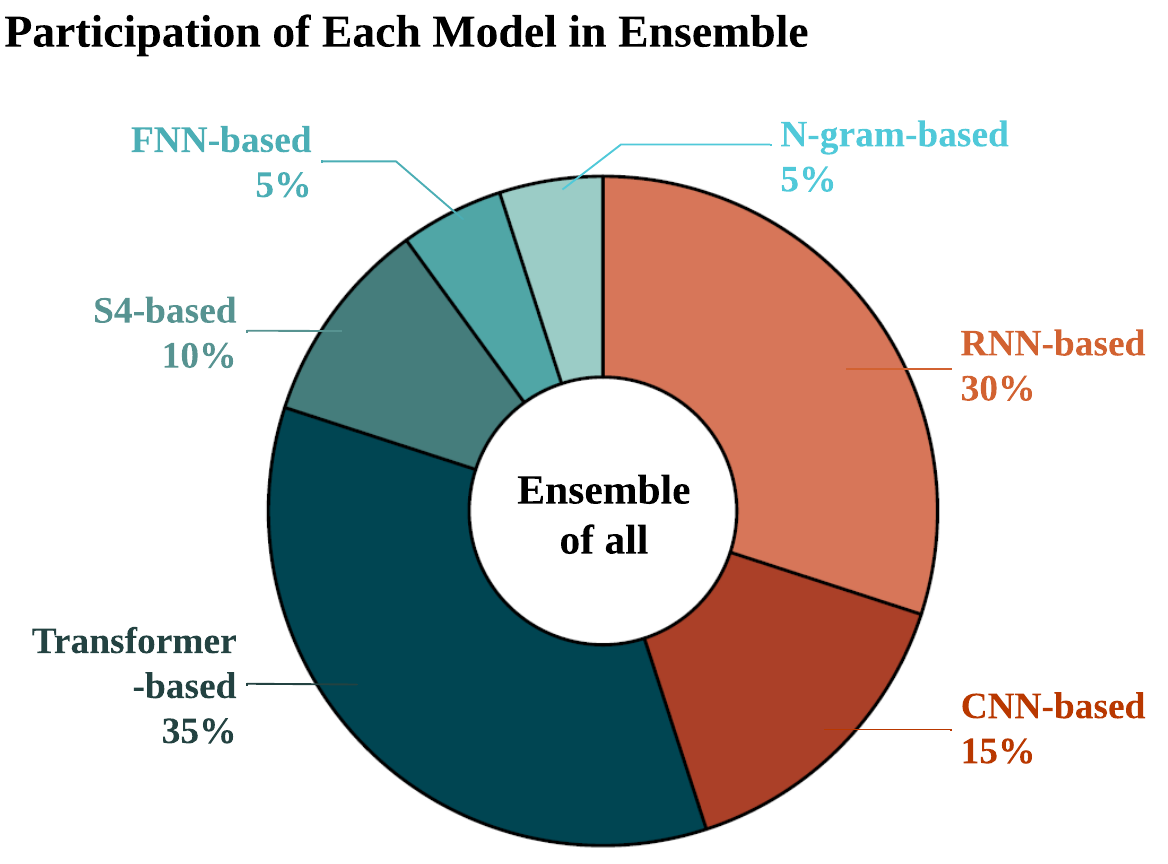}
\caption{This visualization illustrates the concept of ensemble application on a language modeling dataset. It enables easy identification of models that are either obsolete or complementary to the current state-of-the-art. Notably, the models do not necessarily have to be state-of-the-art themselves; they simply need to complement the existing state-of-the-art models effectively.}
\label{fig:ensemble}
\end{figure}

This has changed and neural language models are the mainstream techniques now. 
However, despite the almost exponential growth of interest in neural language modeling in the last decade, it is still very difficult to advance the state of the art. Many top language models today are simply too large and expensive to be explored efficiently in academia, especially their training phase. While there is a range of benchmarks that are more suitable for research purposes due to the limited training data, it is still very hard to reproduce the state-of-the-art results. Many projects are published with open-source code that doesn't work as advertised in the papers, or with rather complex implementations that are very difficult to extend. Since the best published results often use several tricks together, any new idea faces a significant barrier which may prevent its publication: it may simply be too hard to show its value and impress the reviewers.

In this paper, we propose to address this issue. We suggest that researchers should publish not just their papers with code, but also with probabilities per valid and test sets. It is then straightforward to see if a newly proposed model is complementary to the current state of the art by simply adding it into an ensemble using a linear combination. The weights of all models in the ensemble would be optimized on the valid set.

Following this approach, we report new state-of-the-art results on the Penn TreeBank, WikiText-2 and Wikitext-103 datasets. We find that some of the best neural language modeling architectures are in fact complementary, mainly the LSTMs and Transformers. Thus, the findings in this paper may convince the research community that there is value in diversity of ideas: we do not need to search for the best single model, but rather use a combination of complementary models. At the same time, our approach can be elegantly used to detect ideas that are simply reinventing a new name for already known tricks, and thus save time for researchers who are mainly interested in advancing the best possible results.

This work brings several key contributions:
\begin{enumerate}
    \item \textbf{Simplified Ensemble Integration}: Our framework has the potential to significantly advance statistical language modeling, simplifying the integration of new models into an ensemble and accelerating scientific progress.
    \item \textbf{New State-of-the-art}: Our ensemble approach has yielded up to a 10\% improvement on various language modeling benchmarks, underscoring the value of model diversity.
    \item \textbf{Open-Sourced Models}: We make all our reproduced results available\footnote{\url{https://github.com/DavidHerel/sota_lm}}, enabling straightforward testing of new models to assess their compatibility with the state-of-the-art. We shift the focus from producing identical state-of-the-art models to introducing complementary models that enhance ensemble performance.
    \item \textbf{Promoting Ensemble Approaches over Single Models}: The scientific community has traditionally prioritized developing single models, overlooking the true potential of ensemble state-of-the-art. This was primarily due to the difficulty in reproducing results, hindering the creation of ensembles. However, our work has transformed this dynamic, enabling the community to progress more rapidly and improve the true state-of-the-art, beyond just individual model performance.
\end{enumerate}

\section{Related work}
Statistical language modeling, a domain that can trace its origins back to the seminal work of Claude Shannon in 1950, has seen a remarkable evolution over the decades. Shannon's work laid the foundation for the use of probabilistic models in predicting the likelihood of a given sequence of words in a language, setting the stage for the development of n-gram models \citep{shannon1951prediction}. For a significant period, these n-gram techniques, which predict the next word in a sequence based on the previous 'n' words, were the gold standard in language modeling. They were used extensively in a variety of applications, from speech recognition to machine translation.

The rise of neural language models marked a pivotal shift. These models, which use neural networks to predict the next word in a sequence, challenged the long-standing belief that advanced alternatives to n-gram models were only effective on smaller datasets, and their performance would decline as the size of the data increased \citep{goodman}. 

Today, the scenario has evolved, and neural language models have become the standard techniques. This shift can be credited to several pivotal factors:
\begin{itemize}
    \item The availability of open-source code, with the first neural language modeling toolkit published in 2010 \cite{penn}.
    \item The presence of open benchmarks, including the Modified Penn TreeBank benchmark from 2010 in the RNNLM project \citep{penn}, as well as datasets like Wikitext-103 \cite{penn}.
    \item Advances in algorithms and hyper-parameter tuning, such as effectively handling large vocabularies \cite{transformers}, gradient clipping \cite{mikolov2010recurent}, utilization of large models with dropouts \cite{transformerxl, gal2016theoretically}.
    \item Increased computational power.
\end{itemize}

However, the community subsequently shifted towards evaluating single models on these datasets. This raises the question of why we should limit ourselves to single models, and what precisely constitutes a single model. Even a large language model with dropouts can be viewed as an ensemble of sorts. To remain relevant, researchers must continually produce new state-of-the-art models, as this is the only way to persuade others. We propose publishing valid and test word probabilities along with code and paper, enabling easy integration into ensembles to determine whether a model introduces innovations to the state-of-the-art or merely repackages existing techniques.

Ensemble learning, a widely adopted technique for enhancing model performance, involves harnessing the strengths of multiple weaker models~\cite{Sagi2018EnsembleLA,Anio2019EnsembleAF}. Typically, ensemble learning entails considering model weights or combining diverse outputs, much like our approach of combining word probabilities. Mix-of-Experts (MoE) represents a specific ensemble approach that combines predictions from multiple specialized sub-models to improve overall performance. MoE has found success in diverse domains, including natural language processing and computer vision~\cite{Jacobs1991AdaptiveMO,Shazeer2017OutrageouslyLN}. However, we demonstrate that it's possible to achieve state-of-the-art results through a simple linear combination of models.

\section{Experiments}
In today's competitive landscape, a model must significantly surpass the prevailing top-performing models to gain recognition. However, when we delve into ensemble setups, the dynamics change. Here, a model may not shine individually, but its unique functionality can bolster the overall ensemble's performance.

To determine which models enhance language modeling results, we've executed rigorous tests on the three most distinguished language modeling datasets. These datasets span from 1 million to 103 million words, as illustrated in Figure \ref{fig:ensemble}. This broad spectrum allows us to illustrate how an increase in data volume can shift model partitioning.

Subsequently, we re-trained and re-evaluated all the models evaluated on these datasets and integrated them into a robust ensemble.

\subsection{Datasets}

\begin{figure}[h]
\centering
\includegraphics[width=0.8\linewidth]{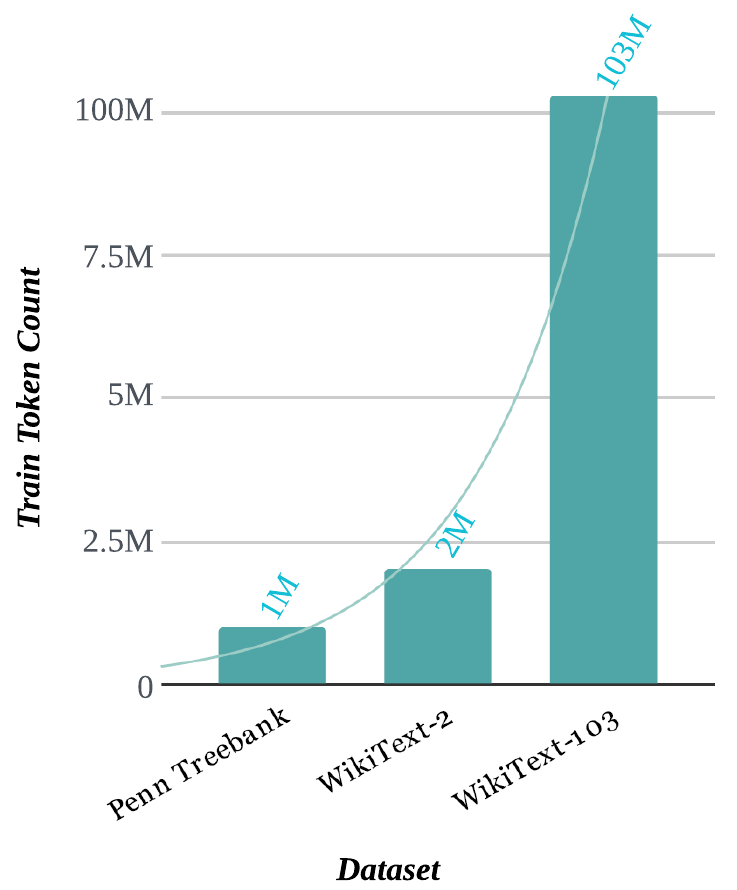}
\caption{This graph illustrates the number of training tokens present in each dataset, specifically 1M, 2M, and 103M \cite{penn, merity2016pointer}. It provides a visual representation of the increase in token numbers across the datasets, enabling an analysis of how the application of different language model architectures evolves in response to increasing data volume.}
\label{fig:ensemble}
\end{figure}

Experiments are conducted on widely accepted language modeling benchmarks. Namely, Penn TreeBank \cite{penn}, Wikitext-2\cite{merity2016pointer} and Wikitext-103\cite{merity2016pointer}. To understand the impact of increased data volume on model partitioning, models are evaluated across datasets ranging from 1 million to 103 million words.

\begin{itemize}
    \item \textsc{Penn TreeBank}~\cite{penn} is a word-level language modeling dataset comprised of Wall Street Journal articles. It stands as one of the most recognized and frequently utilized corpora for evaluating language modeling models. The Penn TreeBank Corpus was segmented as follows: sections 0-20 served as the training data with 930k tokens, sections 21-22 were designated as the validation data with 74k tokens, and sections 23-24 were allocated as the test data with 82k tokens. All words beyond the 10K vocabulary were assigned to a unique token (unknown word) across all Penn TreeBank datasets, thereby eliminating Out-Of-Vocabulary (OOV) words.
    
    \item \textsc{WikiText-2}~\cite{merity2016pointer} is a robust word-level language modeling dataset. This dataset, derived from validated Good and Featured articles on Wikipedia, is more than twice the size of the preprocessed Penn TreeBank (PTB) dataset. Additionally, it offers a significantly larger vocabulary and maintains the original case, punctuation, and numbers - elements that are entirely removed in PTB. Given its comprehensive nature, consisting of full-length articles, the dataset is perfectly tailored for models designed to capitalize on long-term dependencies.
    
    \item \textsc{WikiText-103}~\cite{merity2016pointer} surpasses the original Penn TreeBank by being a hundred times larger. With the same foundation in validated Good and Featured articles on Wikipedia and identical preprocessing cleaning as in WikiText-2, it brings a breadth of 103 million training words and a vocabulary spanning 267k words.

\end{itemize}

\subsection{Models}
In our efforts to build a comprehensive ensemble of models, we've included a wide variety of models for each dataset. This includes Transformers, Recurrent Neural Networks (RNNs), Feed-Forward Neural Networks (FNNs), Convolutional Neural Networks (CNNs), N-gram based models and others.

We've gone through the process of retraining and reproducing every possible model that had previously been evaluated on these datasets. These models were then added to our ensemble. Our goal in doing this was to determine which models genuinely add value to the state-of-the-art and which ones don't contribute anything new and are therefore not useful in the larger context.

\subsubsection{Transformer-based models}
In terms of models based on the Transformer architecture \cite{transformers}, we've selected a variety of options. First, we've chosen Transformer-XL \cite{transformerxl}, a model that facilitates learning dependencies that extend beyond a fixed length. We've also incorporated ShortFormer \cite{press-etal-2021-shortformer}, a model designed to work effectively with shorter inputs. Additionally, we've included Mega model \cite{ma2023mega}, a Transformer-based model that utilizes a more efficient attention mechanism. Finally, we've chosen the DeLight model \cite{delight}, notable for its performance despite using significantly fewer parameters than a standard Transformer model.

\subsubsection{RNN-based models}
Regarding Recurrent Neural Networks (RNNs) \cite{Elman90}, we've utilized a variety of models. We've made use of an LSTM-based \cite{lstm} network with several modifications, popularly known as AWD-LSTM \cite{awd-lstm}. Additionally, we've incorporated an LSTM model equipped with associative memory, known as FWM \cite{fwm}. We've also included models that are based on GRU \cite{gru}, specifically the EGRU model \cite{egru}. The EGRU model is inspired by biological neuron dynamics, leading to sparse and discrete communication between RNN units. This results in a backward pass with backpropagation through time (BPTT) that is both computationally sparse and efficient.

\subsubsection{CNN-based models}
Our ensemble includes models based on Convolutional Neural Networks (CNN). We used QRNN \cite{qrnn}, a neural sequence modeling method that alternates between convolutional layers, which work simultaneously across different time steps, and a basic recurrent pooling function that works simultaneously across different channels. We also used Gated Convolutional Networks (GCN) \cite{gcn} that create a limited context approach through stacked convolutions. This approach is more efficient because it allows for parallel processing over sequential tokens.

\subsubsection{N-gram Models}
We selected a 5-gram model with KneserNey smoothing \cite{kneserney} for our N-gram model. This smoothing method has consistently shown top performance and is widely used.

\subsubsection{Additional Architectures}
We incorporated a new architecture known as Structured State Spaces (S4) \cite{s4}, which outperforms specific Transformers architecture. We also integrated a method called Nearest Neighbor Language Models \cite{knnlm}. In this approach, the language model is combined linearly with a k-nearest neighbor model.

\subsubsection{Exclusion of Dynamic Evaluation and Cache Models}

In the development of our ensemble, we've consciously opted not to include cache models or those utilizing dynamic evaluation. This decision stems from the fact that not all models we reproduced have dynamic evaluation implemented. Adding this feature ourselves would present significant challenges, given the numerous unique codebases we would have to modify. This could introduce potential inconsistencies or errors and result in sub-optimal outcomes.

Moreover, our focus lies in enhancing existing models within their original design constraints. Incorporating dynamic evaluation deviates from the models' original design and exceeds the scope of our current project. By excluding dynamic evaluation and cache models, we ensure a consistent approach to our ensemble creation.

\subsection{Linear combination of models} 
Our approach to ensemble modeling involves a strategic linear combination of multiple models to enhance the accuracy of predictions. We employ a method of soft voting, which involves a weighted average of the predicted word probabilities. This process enables us to integrate the strengths of individual models, striking a balance that leverages the unique capabilities of each model.

In this model, individual predictions, denoted by $P_{j}(w_{i}|w_{1}...w_{i-1})$, are weighted by a specific weight, $m_{j}$, which is a positive value and sums up to one. This is used to calculate the final probability of a word, $w_{i}$, following the formula:
$$P(w_{i}|w_{1}...w_{i-1}) = \sum_{j=0}^{K}m_{j}P_{j}(w_{i}|w_{1}...w_{i-1})$$
where $K$ represents the total number of ensemble models.

The determination of weights, $m_{j}$, is a critical step in our process. We achieve this by minimizing the cross-entropy, $H(W)$, on the validation set, following the formula:
$$H(W) = -\frac{1}{N}\sum_{i=1}^{N}\log(P(w_{i}|w_{1}...w_{i-1}))$$
where $N$ is the total number of words. Here, the softmax function is applied to ensure that the weights, $m_{j}$, are positive and that their total sum is equal to one.

By minimizing cross entropy, we are directly minimizing the perplexity, a measure of how well a probability distribution or probability model predicts a sample, as defined by \cite{Jurafsky:2009:SLP:1214993}:
$$PP(W) = e^{H(W)}$$ 
This process allows us to evaluate the performance of our ensemble model and iteratively improve the weighting and combination of models to achieve optimal results.

\section{Results}

\subsection{Penn TreeBank Results}

\renewcommand{\arraystretch}{1.2}
\begin{table*}[tp]
\centering
\newcommand\mycolsize{0.075\linewidth}
\newcommand\mycolsizebis{0.1\linewidth}
\newcommand\mycolsizetime{0.055\linewidth}
{\scriptsize
\begin{tabular}{p{0.31\linewidth}|p{\mycolsize}|p{\mycolsizebis}|p{\mycolsizetime}p{-20pt}p{\mycolsize}|p{\mycolsizebis}|p{\mycolsizetime}}
 \multicolumn{4}{c}{\normalsize (a) \textbf{Penn TreeBank}}&&\multicolumn{3}{c}{\normalsize (b) \textbf{WikiText-2}}\\
\cmidrule[\heavyrulewidth]{1-4} \cmidrule[\heavyrulewidth]{6-8}
\bfseries \small Model & \bfseries \small Weight $\uparrow$ & \bfseries \small Validation $\downarrow$ & \bfseries \small Test $\downarrow$ & & \bfseries \small Weight $\uparrow$ & \bfseries \small Validation $\downarrow$ & \bfseries \small Test $\downarrow$ \\
\cmidrule{1-4} \cmidrule{6-8}
\small QRNN \cite{qrnn} & \small  0.00 & \small 60.38 & \small 58.43 & & \small 0.00 & \small 69.23 & \small 66.61\\
\small KnerserNey-5gram \cite{kneserney} & \small  0.01 & \small 148.41 & \small 141.46 & & \small 0.01 & \small 233.93 & \small 219.27\\
\small Fast Weights \cite{fwm} & \small  0.11 & \small 58.49 & \small 56.45 & & \small 0.03& \small 69.51 & \small 66.49\\
\small EGRU \cite{egru} & \small  0.15 & \small 61.21 & \small 57.18 & & \small 0.22& \small 69.40 & \small 67.20\\
\small AWD-LSTM-MOS \cite{awd-lstm-mos} & \small  0.19 & \small 56.04 & \small 54.00 & & \small 0.17 & \small 63.92 & \small 61.54\\
\small Transformer-XL \cite{transformerxl} & \small  0.20 & \small 57.93 & \small 55.41 & & \small 0.16 & \small 67.41 & \small 64.85\\
\small AWD-LSTM-DOC \cite{awd-lstm-doc} & \small  0.34 &  \small 54.12 & \small 52.38 & & \small 0.41 & \small 60.27 & \small 58.01\\
\cmidrule[\lightrulewidth]{1-4} \cmidrule[\lightrulewidth]{6-8}
\addlinespace[+\aboverulesep]
\textbf{\small \textit{Ensemble of All}} & \small  1 & \bfseries \small 48.92& \bfseries \small 47.31 & &\small 1& \bfseries \small 55.40 &\bfseries \small 53.73\\
\cmidrule[\lightrulewidth]{1-4} \cmidrule[\lightrulewidth]{6-8}
\addlinespace[+\aboverulesep]
\end{tabular}
}
\caption{Perplexity on validation and test sets for Penn TreeBank and WikiText-2 dataset \cite{penn, merity2016pointer}. First column corresponds to the model name. Second column refers to the weight the model receives in the ensemble. Valid and test refers to the obtained perplexity on validation and test set. A downward arrow ($\downarrow$) indicates that lower is better.
}
\label{table:ptb_wiki}
\vspace{5pt}
\end{table*}
\renewcommand{\arraystretch}{1}

\begin{table*}[ht!]
\small
\centering
\begin{tabular}{p{0.33\linewidth}|l|l|l}
\multicolumn{4}{c}{\normalsize (c) \textbf{WikiText-103}}\\
\toprule
\textbf{Model} & \textbf{Weight $\uparrow$} & \textbf{Validation $\downarrow$} & \textbf{Test $\downarrow$}\\
\midrule
Delight \cite{delight}& 0.00 & 42.77 & 44.42\\
GCN \cite{gcn}& 0.00 & 34.34 & 35.81\\
AdaptiveInputs \cite{adaptive-inputs}& 0.03 & 18.15 & 18.86\\
S4 \cite{s4}& 0.06 & 20.47 & 21.14\\
ShortFormer \cite{press-etal-2021-shortformer}& 0.09 & 17.47 & 18.15\\
MEGA \cite{ma2023mega}& 0.29 & 17.17 & 18.09\\
kNN LM \cite{knnlm}& 0.53 & 14.39 & 14.54\\
\midrule
\textbf{\textit{Ensemble of All}} & 1 & \textbf{13.11} & \textbf{13.29} \\
\bottomrule
\end{tabular}
\caption{\label{results_wt103}
Perplexity on validation and test sets for WikiText-103 (WT-103) dataset \cite{merity2016pointer}. First column corresponds to the model name. Second column refers to the weight the model receives in the ensemble. Valid and test refers to the obtained perplexity on validation and test set. A downward arrow ($\downarrow$) indicates that lower is better. Since previous model authors did not train and evaluate their models on larger datasets like WT-103, we had to identify models that were, leading to the creation of the ensemble with different models.
}
\end{table*}

Our analysis of the Penn TreeBank dataset from Table \ref{table:ptb_wiki} reveals several intriguing findings. Firstly, we note that the QRNN model \cite{qrnn}, despite achieving a perplexity of 58.43 on the test data, does not contribute any weight to our ensemble. In contrast, the KneserNey 5-gram model \cite{kneserney}, which exhibits perplexity more than twice as high, is assigned a higher weight in the ensemble. This discrepancy can be attributed to the distinct behaviors of these models; the 5-gram model operates differently from the QRNN, which is constructed on neural network foundations combining RNN and CNN architectures. Since these architectural components are already represented in our ensemble, it appears that the QRNN does not provide any additional benefits.

Furthermore, we observe that models based on Transformers \cite{transformers} and RNN \cite{Elman90} architectures dominate the ensemble's weights, indicating their superior performance in this small-scale language modeling task. Interestingly, even though the AWD-LSTM-MOS \cite{awd-lstm-mos} model exhibits better perplexity scores compared to Transformer-XL \cite{transformerxl}, the latter is assigned a higher weight in the ensemble. This disparity can be attributed to the complementarity of Transformer-XL within the ensemble, as it demonstrates a distinct behavior from the LSTM-based models.

Incorporating these diverse models into our ensemble leads to a substantial improvement in perplexity, surpassing the state-of-the-art AWD-LSTM-DOC \cite{awd-lstm-doc} by over 10\%.

\subsection{WikiText-2 Results}

Results obtained on the Wiki-Text 2 dataset, which is two times larger than the previously used Penn TreeBank dataset, highlight the prominence of the AWD-LSTM-DOC model \cite{awd-lstm-doc}, followed closely by EGRU \cite{egru}, a distinct variant of LSTM \cite{lstm}, as we can see in Table \ref{table:ptb_wiki}.

Interestingly, despite KnerserNey exhibiting more than a threefold higher perplexity compared to QRNN, KneserNey has a much higher contribution to our ensemble whereas QRNN receives 0 weight. This phenomenon mirrors our observations on the Penn TreeBank dataset, where n-gram models exhibit distinctly different behavior from neural networks. Consequently, they retain their significance within the ensemble, despite their significantly lower performance.

The combination of these diverse models in our ensemble leads to a substantial improvement in perplexity, surpassing the state-of-the-art AWD-LSTM-DOC \cite{awd-lstm-doc} model by more than 8\%.

\subsection{Wikitext-103 Results}

\begin{figure*}[h]
\centering
\includegraphics[width=0.9\linewidth]{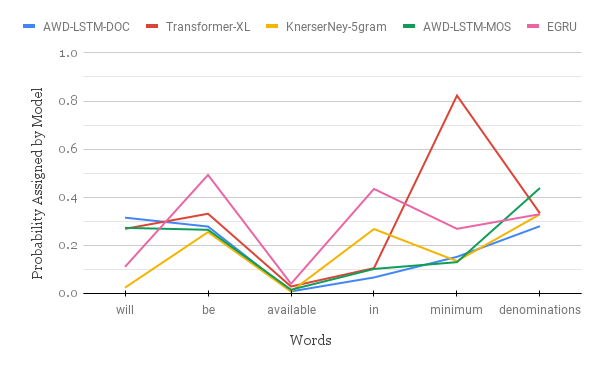}
\caption{This figure illustrates the complementary nature of the models within the ensemble. It showcases the predicted probabilities for each word in the sentence \textit{'will be available in minimum denominations'} as determined by each model in the ensemble. Notably, one of the RNN-based models, EGRU \cite{egru}, surpasses other models with its predictions for the words \textit{'be'} and \textit{'in'}, by roughly \textit{0.2}. On the other hand, the Transformer-XL model \cite{transformerxl} surpasses other models predictions for the word \textit{'minimum'} by more than \textit{0.5}. This underscores the value of having diverse models in the ensemble, justifying why both Transformer-XL and EGRU are assigned substantial weights of \textit{0.2} and \textit{0.15} respectively in the ensemble on the Penn TreeBank test set.}
\label{fig:prob_sent1}
\end{figure*}

Wikitext-3 represents the largest dataset from which we were able to assemble a substantial ensemble of models. With its dataset size being 100 times larger than the Penn TreeBank, our findings on this corpus reveal intriguing insights, as can be seen in Table \ref{results_wt103}.

As most previous models were not trained or evaluated on larger datasets like WT-103, we sought models that were trained specifically on this dataset. Due to the complexity of finding accurate hyperparameters across various codebases, we chose to train and evaluate only those models with provided hyperparameters, enabling us to replicate the results.

Remarkably, the kNN LM \cite{knnlm} model emerges as the dominant contributor, receiving more than half of the ensemble's weight. This underscores the value of leveraging kNN neighbors in language modeling, demonstrating its complementarity with other model types.

Following closely is MEGA \cite{ma2023mega}, which receives nearly 0.3 of the ensemble weight. This result suggests that an alternative attention mechanism, distinct from the conventional approach, plays a complementary role in our ensemble. Additionally, our exploration of newer architectures, such as S4 \cite{s4}, demonstrates that it still contributes with a weight of 0.06, although Transformer-based models significantly outweigh it.

In contrast, models based on CNN, specifically GCN \cite{gcn}, and the Delight model \cite{gcn}, do not perform well in this context, receiving zero weight in our ensemble.

With a substantial reduction in perplexity of over 9\%, our ensemble of the best models has achieved significant improvements compared to the state-of-the-art kNN LM \cite{knnlm}.

\subsection{Complementarity in Model Ensemble}

This section offers practical examples that highlight how different models within the ensemble complement each other. We conducted a series of tests on the Penn TreeBank test set \cite{penn} to find out the biggest differences in word probabilities between the models. These findings shine a light on the strength of using an ensemble approach. In situations where one model might have difficulty, another model steps up to give a more precise prediction, boosting the overall precision of the ensemble. A specific example of this phenomenon on a sample sentence is presented in Figure \ref{fig:prob_sent1}.

Figure \ref{fig:prob_sent1} displays how the models in the ensemble work together. It presents the predicted probabilities for each word in the sentence 'will be available in minimum denominations' as determined by each model in the ensemble. Interestingly, EGRU model \cite{egru} does better than others in predicting the words 'be' and 'in' by about 0.2. On the other hand, the Transformer-XL model \cite{transformerxl} is better at predicting the word 'minimum' by over 0.5. This shows why it is beneficial to have different types of models in the ensemble and why both Transformer-XL and EGRU have larger weights in the ensemble on the Penn TreeBank test set. The image also shows that the predictions between models can change for each word, with some increasing and some decreasing, e.g. for the word 'minimum'.

In an effort to provide a more comprehensive understanding of the complementarity of the models in the ensemble, further examples and detailed explanations are available in Appendix \ref{sec:appendix}. These additional examples not only support the findings presented in this section but also provide deeper insights into the dynamic interplay between different models in the ensemble.
\newpage
\section{Conclusion}
In conclusion, our framework has significantly advanced the state of the art in statistical language modeling. By advocating for the publication of code and probabilities on development and test sets, we have simplified the integration of new models into an ensemble. This has led to a more efficient determination of a model's complementarity to the current state of the art, thereby accelerating scientific progress.

Our approach has yielded remarkable results, with up to a 10\% improvement on various language modeling benchmarks, including the Penn TreeBank, WikiText-2, and Wikitext-103 datasets. This underscores the value of model diversity and the potential contribution of suboptimal models when combined effectively.

Crucially, our approach has also addressed a longstanding issue within the scientific community: the traditional focus on developing single models, which often overlooks the immense potential of ensemble state-of-the-art. This was primarily due to the challenges in reproducing results, which impeded the creation of ensembles. Our work has radically transformed this dynamic, enabling the community to progress more rapidly and enhance the true state-of-the-art, beyond just individual model performance.

We have made all our reproduced results on these datasets available, allowing for easy testing of new models to determine their true complementarity to the state-of-the-art\footnote{\url{https://github.com/DavidHerel/sota_lm}}. It is no longer necessary to produce a new state-of-the-art model that behaves identically to others; instead, it is sufficient to produce a complementary model, which despite potentially underperforming on its own still enhances the ensemble's performance.

We believe that our approach, promoting transparency and reproducibility, can revolutionize the field by lowering barriers for new ideas and eliminating redundant efforts. We encourage the research community to adopt this framework for more rapid and diverse advancements in language modeling.

\section*{Limitations}
While our approach has demonstrated significant advancements, it is important to acknowledge certain limitations. Firstly, the use of ensembles inherently increases the complexity of the code base. This increased complexity could potentially make it harder to maintain the code.

Our models were trained on data up to 100M words, due to the computational power available to us. This, however, does not limit the applicability of our approach to larger datasets. We encourage future work, especially by groups with greater computational resources, to apply our framework to larger benchmarks. 

\section*{Acknowledgement}
We would like to thank Daniela Hradilova for the useful discussions and suggestions. Our work is part of the RICAIP project that has received funding from the European Union’s Horizon 2020 research and innovation programme under grant agreement No. 857306.

\bibliography{custom}
\newpage
\appendix
\section{Detailed Analysis of Model Interplay in Ensemble}
\label{sec:appendix}

This Appendix section offers practical examples that highlight how different models within the ensemble complement each other. We conducted a series of tests on the Penn TreeBank test set \cite{penn} to find out the biggest differences in word probabilities between the models. These findings shine a light on the strength of using an ensemble approach. In situations where one model might have difficulty, another model steps up to give a more precise prediction, boosting the overall trustworthiness of the ensemble.

\begin{minipage}[c]{\textwidth}
\vspace{30pt}
\centering
\includegraphics[width=0.9\linewidth]{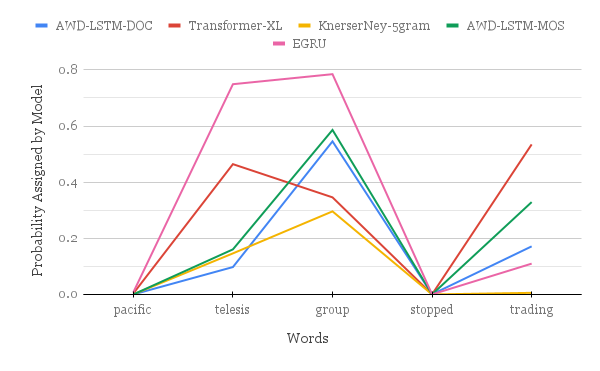}
\captionof{figure}{The figure presents the predicted probabilities for each word in the sentence \textit{’pacific telesis group stopped trading’} as determined by each model in the ensemble on the Penn TreeBank test set. The complementarity of the models within the ensemble is evident. For instance, the EGRU \cite{egru} model significantly outperforms other models in predicting the words \textit{'telesis'} and \textit{'group'}, while the Transformer-XL \cite{transformerxl} excels in predicting the word \textit{'trading'}.}
\label{fig:prob_sent2}
\end{minipage}

\begin{figure*}[ht!]
\centering
\includegraphics[width=0.9\linewidth]{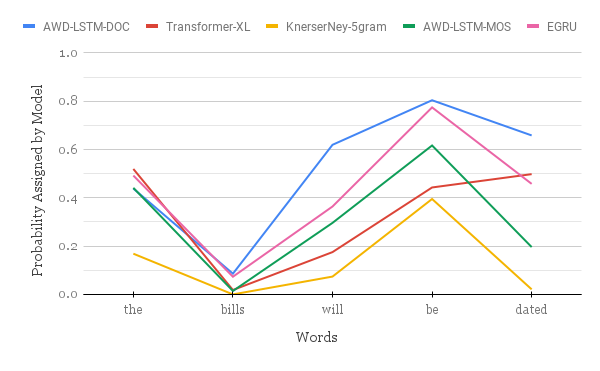}
\caption{This figure depicts the predicted probabilities for the sentence \textit{'the bills will be dated'} as determined by each model in the ensemble on the Penn TreeBank test set. Despite the general similarity in predictions among the models, the Transformer-XL \cite{transformerxl} model displays a unique behavior in predicting the word \textit{'dated'}, demonstrating an increased probability compared to the other models.}
\label{fig:prob_sent3}
\end{figure*}

\end{document}